\documentclass[11pt,letterpaper]{article}
\usepackage[hyperref]{naaclhlt2019}
\usepackage{times}
\usepackage{latexsym}
\usepackage{amsmath}
\usepackage{multirow}
\usepackage{url}
\usepackage{amsthm,amsmath,amssymb,graphicx}
\usepackage{algorithm2e}
\usepackage{algpseudocode}

\makeatletter
\newcommand{\@emptybiblabel}[1]{}
\makeatother

\usepackage{amsfonts}
\usepackage{graphicx}
\usepackage{tabularx}

\usepackage{booktabs} % To thicken table lines
\usepackage{caption}
\usepackage{subcaption}

\usepackage{enumitem}
\setenumerate{leftmargin=5.5mm,itemsep=0pt,topsep=3pt}

\usepackage{tikz}
\usetikzlibrary{positioning}
\usetikzlibrary{shapes.geometric, arrows}
\tikzstyle{startstop} = [rectangle, rounded corners, minimum width=1cm, minimum height=0.5cm,text centered, draw=black]
\tikzstyle{io} = [trapezium, trapezium left angle=70, trapezium right angle=110, minimum width=1cm, minimum height=0.5cm, text centered, draw=black]
\tikzstyle{process} = [rectangle, minimum width=1cm, minimum height=0.5cm, text centered, draw=black]
\tikzstyle{decision} = [diamond, inner sep=0.05cm, aspect=2, text centered, draw=black]
\tikzstyle{arrow} = [thick,->,>=stealth]

\newcommand{\eat}[1]{\ignorespaces}

\newcommand{\commentout}[1]{}

\newcommand{\gleu}{\textrm{GLEU$^+$}}

\newcommand{\ffive}{\textrm{$F_{0.5}$ }}

\colorlet{gold}{green!10!orange!90!}

\newcommand{\ignore}[1]{}

\aclfinalcopy
\title{Corpora Generation for Grammatical Error Correction}

\author{Jared Lichtarge$^{\ast}$, Chris Alberti$^{\ast}$, Shankar Kumar$^{\ast}$, Noam Shazeer$^{\ast}$, Niki Parmar$^{\ast}$, Simon Tong\thanks{{}$^{\ast}$Equal contribution. Listing order is random. Jared conducted systematic experiments to determine useful variants of the Wikipedia revisions corpus, pre-training and fine-tuning strategies, and iterative decoding. Chris implemented the ensemble and provided background knowledge and resources related to GEC. Shankar ran training and decoding experiments using round-trip translated data. Jared, Chris and Shankar wrote the paper. Noam identified Wikipedia revisions as a source of training data. Noam developed the heuristics for using the full Wikipedia revisions at scale and conducted initial experiments to train Transformer models for GEC. Noam and Niki provided guidance on training Transformer models using the \textit{Tensor2Tensor} toolkit. Simon proposed using round-trip translations as a source for training data, and corrupting them with common errors extracted from Wikipedia revisions. Simon generated such data for this paper.}\\
  Google Research\\
  {\tt \{lichtarge,chrisalberti,shankarkumar,noam,nikip,simon\}@google.com} \\}
\date{}

\begin{document}

\maketitle
\begin{abstract}
 Grammatical Error Correction (GEC) has been recently modeled using the sequence-to-sequence framework. However, unlike sequence transduction problems such as machine translation, GEC suffers from the lack of plentiful parallel data.  We describe two approaches for generating large parallel datasets for GEC using publicly available Wikipedia data. The first method extracts source-target pairs from Wikipedia edit histories with minimal filtration heuristics, while the second method introduces noise into Wikipedia sentences via round-trip translation through bridge languages. Both strategies yield similar sized parallel corpora containing around 4B tokens. We employ an iterative decoding strategy that is tailored to the loosely supervised nature of our constructed corpora. We demonstrate that neural GEC models trained using either type of corpora give similar performance. Fine-tuning these models on the Lang-8 corpus and ensembling allows us to surpass the state of the art on both the CoNLL-2014 benchmark and the JFLEG task. We provide systematic analysis that compares the two approaches to data generation and highlights the effectiveness of ensembling.
\end{abstract}

\section{Introduction}
\label{sec:intro}
Much progress in the Grammatical Error Correction
(GEC) task can be credited to approaching the problem as a translation
task~\cite{brockett2006correcting} from an ungrammatical source language to a grammatical target language. This has enabled Neural Machine Translation (NMT) sequence-to-sequence (S2S) models and techniques to be applied to the GEC task~\cite{ge2018,chollampatt2018multilayer,junczys2018approaching}. However, the efficacy of NMT techniques is degraded for low-resource tasks~\cite{koehn17}. This poses difficulties for S2S approaches to GEC, as Lang-8, the largest publicly available parallel corpus, contains only $\sim$$25$M words~\cite{mizumoto2011mining}. Motivated by this data scarcity, we present two contrasting approaches to generating parallel data for GEC that make use of publicly available English language Wikipedia revision histories\footnote{\url{https://dumps.wikimedia.org/enwiki/latest/}}\footnote{Last accessed: December 15, 2017}.

Our first strategy is to mine real-world errors. We attempt to accumulate source--target pairs from grammatical errors and their human-curated corrections gleaned from the Wikipedia revision histories. Unlike previous work~\cite{grundkiewicz2014wiked}, we apply minimal filtering so as to generate a large and noisy corpus of $\sim$$4$B tokens (Table~\ref{tab:tokens}). As a consequence of such permissive filtering, the generated corpus contains a large number of real grammatical corrections, but also noise from a variety of sources, including edits with drastic semantic changes, imperfect corrections, ignored errors, and Wikipedia spam.

Our second strategy is to synthesize data by corrupting clean sentences. We extract target sentences from Wikipedia, and generate corresponding source sentences by translating the target into another language and back. This round-trip translation introduces relatively clean errors, so the generated corpus is much less noisy than the human-derived Wikipedia corpus. However, these synthetic corruptions, unlike human errors, are limited to the domain of errors that the translation models are prone to making. Both approaches benefit from the broad scope of topics in Wikipedia.

\begin{table}[ht]
  \footnotesize
  \centering
  \begin{tabular}{c|c|c}
    \toprule
    Corpus & \# sentences & \# words \\
    \midrule
    Lang-8 & 1.9M & 25.0M \\
    WikEd & 12M & 292 M \\ \hline
    Wikipedia Revisions & 170M  & 4.1B  \\
    Round-Trip Translation & 176M  & 4.1B \\
    \bottomrule
  \end{tabular}
  \caption{Statistics computed over extant training sets for GEC (top) and corpora generated from Wikipedia in this work (bottom).}
  \label{tab:tokens}
\end{table}

% Both of our data generation schemes are noisy; unlike corpuses manually curated for GEC, we use target sentences gleaned from Wikipedia that are not guaranteed to be free of grammatical errors. Neural models trained on such noisy data learn to make fewer corrections at inference time, particularly if there are multiple corrections to be made in the input sentence. To alleviate this problem, we employ an iterative decoding algorithm that allows for incremental corrections. While prior work has explored iterative algorithms for GEC inference to progressively expand the search space using either a phrase-based machine translation translation~\cite{dahlmeier2012iterative} or with a sequence-to-sequence model with fluency scores~\cite{ge2018b}, we demonstrate that iterative decoding is effective at addressing the restraint of neural models trained on noisy data.

We train the Transformer sequence-to-sequence model~\cite{vaswani2017attention} on data generated from the two schemes. Fine-tuning the models on the Lang-8 corpus gives us additional improvements which allow a single model to surpass the state-of-art on both the CoNLL-2014 and the JFLEG tasks. Finally, we explore how to combine the two data sources by comparing a single model trained on all the data to an ensemble of models.

% The paper is organized as follows: In Section~\ref{sec:wikidata}, we describe the parallel corpus generation scheme for mining Wikipedia revision histories.
% In Section~\ref{sec:roundtrip} we present the corpus generation scheme using round-trip translations.
% We implement an iterative decoding algorithm, described in Section~\ref{sec:decode}.
% In Section~\ref{sec:model} we describe the neural model used throughout this work.
% We report experiments performed in Section~\ref{sec:experiments}.
% An error analysis is provided in Section~\ref{sec:eval_analysis}.
% We discuss related work in Section~\ref{sec:related} and present a discussion in Section~\ref{sec:discussion}.

\section{Data Generation from Wikipedia Revision Histories}
\label{sec:wikidata}

\begin{figure*}[t!]
	\centering
	\includegraphics[width=\textwidth]{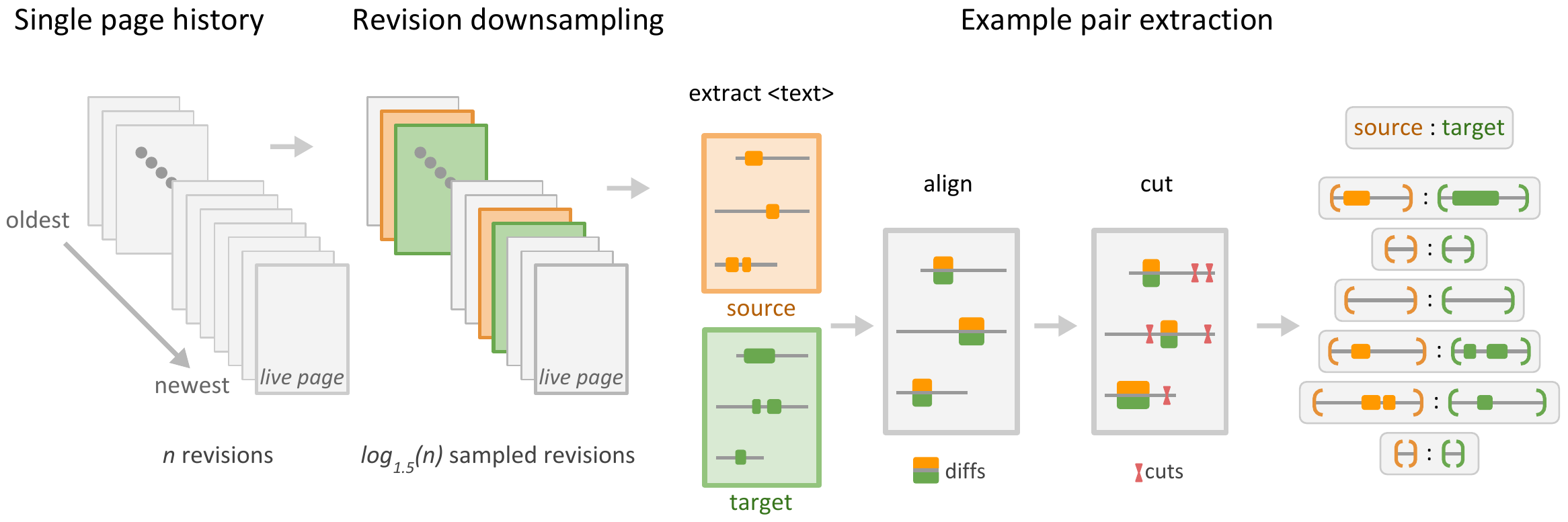}
  \caption{Process for extracting source--target pairs from revision history of a Wikipedia page. See Figure~\ref{fig:wiki_examples} for actual examples.}
	\label{fig:wiki_extraction}
	\vspace{0.5cm}
\end{figure*}

% \subsection{Data Source}
% Wikipedia is a publicly available, online encyclopedia for which all content is
% communally created and curated.
Wikipedia provides a dump of the revision histories of all Wikipedia pages.
For each Wikipedia page, the dump contains chronological snapshots
of the entire content of the page before and after every submitted edit; thus two consecutive snapshots characterize a single revision to the page.
Because a small number of popular pages see disproportionate traffic, some pages grow very large. As we are interested in the edits between snapshots, and not identical content that is typically in higher proportion in the revision histories for the largest pages, we discard pages larger than 64Mb.
To prevent remaining large pages from skewing the dataset towards their topics with their many revisions, we downsample consecutive revisions from individual pages,
selecting only $\log_{1.5}(n)$ pairs for a page with a total of $n$ revisions.
This reduces the total amount of data 20-fold.
Each remaining pair of consecutive snapshots forms a source--target pair. The process for extracting examples from a page's revision history is illustrated in Figure~\ref{fig:wiki_extraction}.

From the XML of each page in a source--target pair, we extract and align the text, removing non-text elements.
We then probabilistically cut the aligned text, skipping over non-aligned sequences.
Two cuts bound an example pair, for which the source sequence is provided by the older snapshot, and the target sequence by the newer snapshot.

Following extraction of the examples, we do a small amount of corruption and filtration
in order to train a model proficient at both spelling and grammar correction. We probabilistically introduce spelling errors in the source sequences at a rate of $0.003$ per
character, randomly selecting deletion, insertion, replacement, or transposition of adjacent characters for each introduced error.

We throw out examples exceeding a maximum length of 256 word-pieces.
The majority of examples extracted by this process have identical source and target. Since this is not ideal for a GEC parallel corpus, we downsample identity examples by 99\% to achieve 3.8\% identical examples in the final dataset.
The data generation scripts we use have been opensourced\footnote{\url{https://github.com/tensorflow/tensor2tensor/blob/master/tensor2tensor/data\_generators/wiki\_revision.py}}.

\begin{figure*}[ht]
  \centering
  \includegraphics[width=\textwidth]{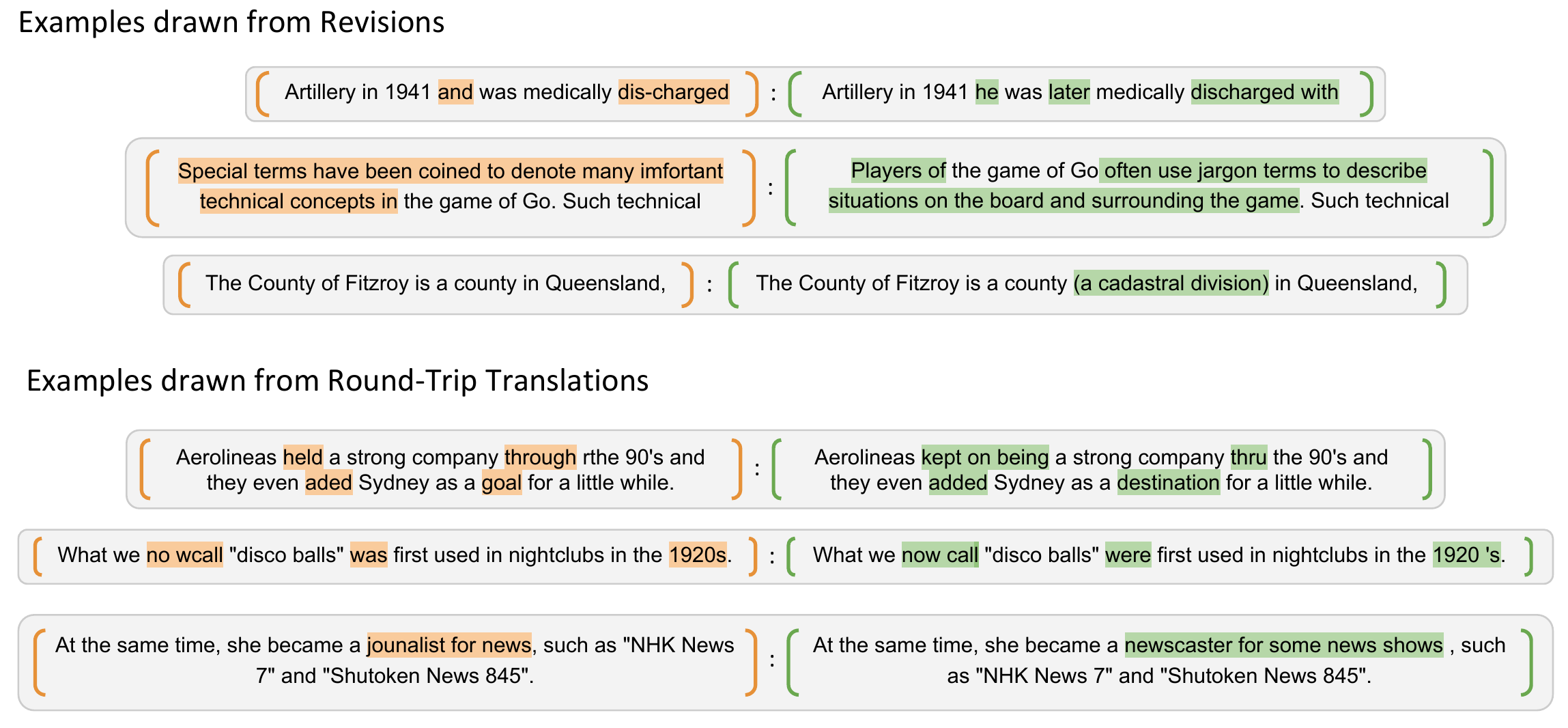}
  \caption{Example source--target pairs from each corpus.}
  \label{fig:wiki_examples}
\end{figure*}

In Figure~\ref{fig:wiki_examples}, we show
examples of extracted source--target pairs. While some of the edits
are grammatical error corrections, the vast majority are not.

\section{Data Generation from Round-trip Translations}
\label{sec:roundtrip}
As an alternative approach to extracting the edits from Wikipedia revisions, we extract sentences from the identity examples that were discarded during edit extraction, and generate a separate parallel corpus by introducing noise into those sentences using round-trip translation via a bridge language. Therefore, the original sentence from Wikipedia is the target sentence and output of the round-trip translation is the corresponding source sentence. The round trip translations introduce noise according to both the weaknesses of the translation models and the various inherent ambiguities of translation. We create a corrupted dataset using each bridge language. We use French (Fr), German (De), Japanese (Ja) and Russian (Ru) as bridge languages because they are high-resource languages and relatively dissimilar from each other. Thus, we compute a total of four corrupted datasets. The translations are obtained using a competitive machine translation system~\cite{googlenmt}.

These round trip translated sentence-pairs contained only a small fraction of identity translations compared to those that are present in real-world GEC corpora. To address this deficiency, we augment this corpus with 2.5\% identity translations. Analogous to Section~\ref{sec:wikidata}, we want the models to learn both spelling and grammar correction. Thus, we randomly corrupt single characters via insertion, deletion, and transposition, each with a probability of $0.005/3$. Round-trip translations do not contain some types of word and phrase errors (e.g., your/you're, should of/should have) and so we additionally corrupt the translated text by stochastically introducing common errors identified in Wikipedia. We first examine the Wikipedia revision histories to extract edits of up to three words whose source and target phrases are close in edit distance, and which do not contain numbers or capitalization. For each of the remaining edits (original, revised), we compute the probability that the user typed \textit{original} when they intended to type \textit{revised}:
\begin{equation*}
  P(\text{original} | \text{revised}) =  \frac{C(\text{original}, \text{revised})}{C(\text{revised})},
\end{equation*}
where $C(x)$ refers to the counts of $x$ in the corpus. We then probabilistically apply these rules to corrupt the translated text.

This process produces a parallel corpus identical in size to the Wikipedia Revision corpus, though with vastly different characteristics. Because the target sentences are Wikipedia sentences that were left unchanged for at least one Wikipedia revision, they are less likely to contain poor grammar, misspellings, or spam than the target sequences of the revisions data.

Also, the errors introduced by round-trip translation are relatively clean, but they represent only a subset of the domain of real-world errors.
In contrast, the Wikipedia data likely has good coverage of the domain of real-world grammatical errors, but is polluted by significant noise. Examples from both corpora are shown in Figure~\ref{fig:wiki_examples}. Examples of round-trip translations for each bridge language are shown in Table~\ref{tab:rtexamples}.

\begin{table*}[ht]
  \footnotesize
  \centering
  \begin{tabular}{c|c}
  % \begin{tabular}{p{1.1cm}|p{5cm}}
    \toprule
    Original & ``The Adventures of Patchhead`` makes its second and final appearance. \\ \midrule
    Bridge Language & \\ \midrule
    French & ``The Adventures of Patchhead `` makes \textbf{his secnod} and final appearance.  \\
    German &  ``The Adventures of Patchhead'' makes its second and \textbf{last} appearance. \\
    Russian & ``The Adventures of Patchhead'' makes its second and \textbf{last apparance}. \\
    Japanese &  \textbf{``Patchhead Adventure'' is the} final appearance \textbf{of the second time}. \\
    \midrule
    Original & He is not so tolerant of the shortcomings of those outside his family. \\ \midrule
    Bridge Language & \\ \midrule
    French & He is not so tolerant of the \textbf{weaknesses} of those outside his family.  \\
    German &  He is not so tolerant \textbf{to} the \textbf{defects} of \textbf{the} outside \textbf{of} his family. \\
    Russian & He is not so tolerant of the shortcomings of those outside his family\textbf{,}. \\
    Japanese &  He is not so tolerant of the shortcomings of those outside his family. \\
    \bottomrule
  \end{tabular}
  \caption{Example sentences generated via round-trip translation with introduced spelling errors.}
  \label{tab:rtexamples}
\end{table*}

\section{Iterative Decoding}
\label{sec:decode}
\begin{figure}
	\centering
	\includegraphics[height=1.6in]{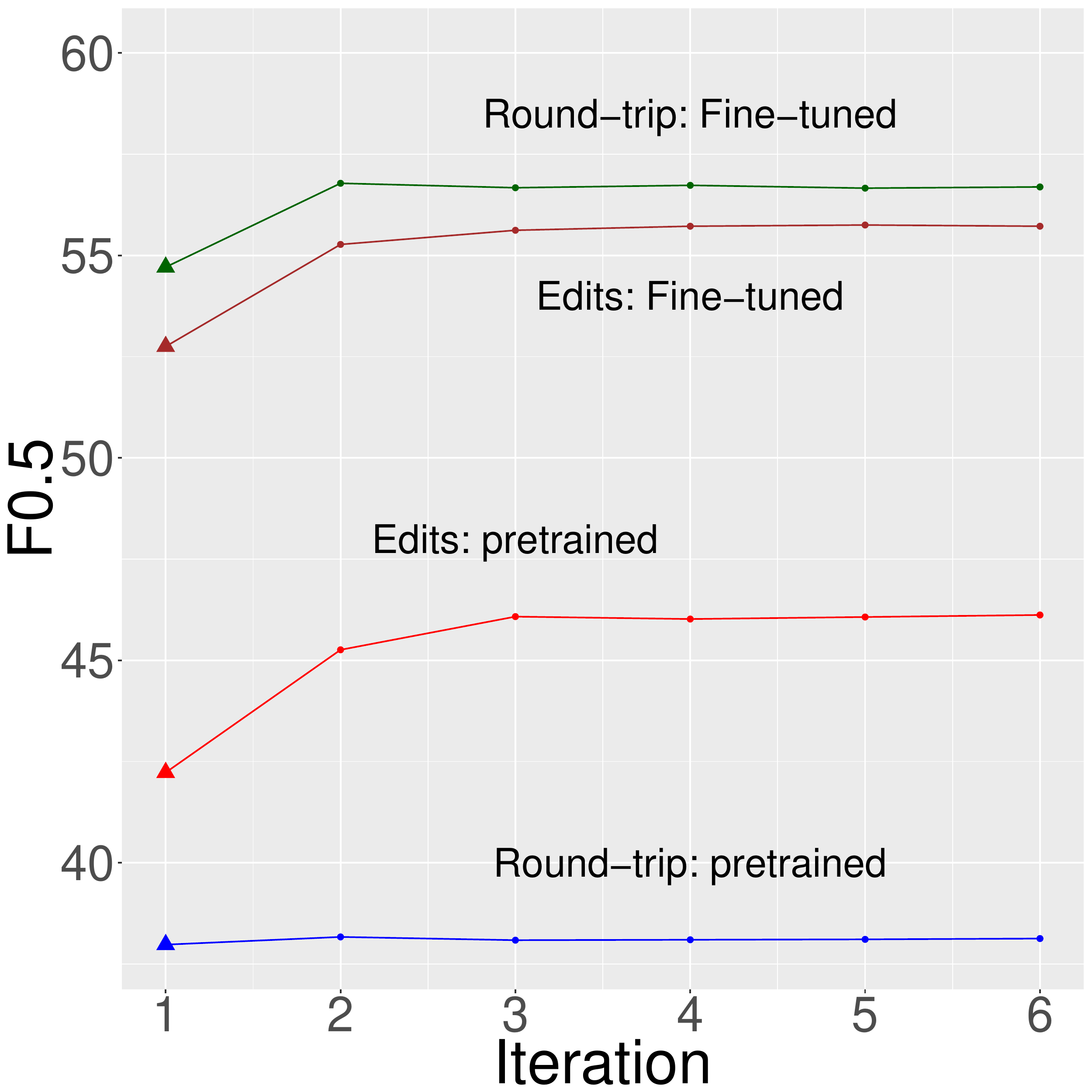}
        \caption{\ffive with iterative decoding on the CoNLL dev set. Triangles indicate performance with single-shot decoding.}
	\label{fig:iterative}
\end{figure}
\begin{table}[ht]
  \centering
  \footnotesize
  \begin{tabular}{cc|ccc|c}
   \toprule
    Source & Decoding & \multicolumn{3}{c}{CoNLL-2014} & JFLEG \\ \hline
    & & Prec. & Rec. & \ffive & \gleu \\ \midrule
    Revision & single-shot & 60.4 & 19.2 & 42.2 & 54.5 \\
     & iterative & 58.3 & 25.1 & 46.1 & 56.6 \\
    +finetune & single-shot & 67.7 & 28.1 & 52.8 & 57.9 \\
     & iterative & 64.5 & 36.2 & 55.8 & 62.0 \\
    \midrule
    RTT & single-shot & 47.1 & 21.4 & 38.0 & 52.5 \\
     & iterative & 47.1 & 21.4 & 38.0 & 52.5 \\
    +finetune & single-shot & 66.7 & 31.8 & 54.7 & 59.0 \\
     & iterative & 64.4 & 38.4 & 56.7 & 62.1 \\
    \bottomrule
  \end{tabular}
  \caption{Comparing iterative decoding to single-shot decoding for two models, trained on all Wikipedia revisions data and on all round-trip translation (RTT) data.}
  \label{tab:singleshotiterative}
\end{table}
\begin{table}[ht]
  \footnotesize
  \centering
  \begin{tabular}{p{1.3cm}|p{5cm}}
    \toprule
    Original & this is nto the pizzza that i ordering \\
    1st & this is \textbf{not} the \textbf{pizza} that \textbf{I} ordering \\
    2nd & \textbf{T}his is not the pizza that I ordering \\
    3nd & This is not the pizza that I \textbf{ordered} \\
    4th & This is not the pizza that I ordered\textbf{.} \\
    Final & This is not the pizza that I ordered. \\
    \bottomrule
  \end{tabular}
  \caption{Iterative decoding on a sample sentence. }
  \label{tab:decode}
\end{table}

Many sentences that require grammatical correction contain multiple errors. As a result, it can be difficult to correct all errors in a single decoding pass. This is specifically a problem when using models trained on noisy parallel data such as Lang-8 where the target sentences still contain grammatical errors. Following other work on the GEC task~\cite{dahlmeier2012iterative,ge2018b}, we employ an iterative decoding algorithm that allows the model to make multiple incremental corrections. This allows the model multiple chances to suggest individually high-confidence changes, accruing incremental improvements until it cannot find any more edits to make.
% We use iterative decoding to address a domain mismatch between the presented Wikipedia-derived corpora and corpora curated for the GEC task.
% Unlike supervised bitext such as CoNLL, both Wikipedia-derived corpora contain target sequences
% that cannot be guaranteed to be free of grammatical errors.
% The models trained on the generated data therefore learn it is
% occasionally acceptable to leave errors in the sequence, particularly if there
% are multiple errors to correct in the input sequence.
% Iterative decoding alleviates this problem by giving the model multiple chances to revise the sequence, making incremental
% improvements until it cannot find any more edits to make.

Our iterative decoding algorithm is presented in
Algorithm~\ref{algo:decode}. Given the source sentence $S$ and a hypothesis $H$, $\text{Cost}(H)$ refers to the negative log probability
$-log P(H|S)$ using the sequence-to-sequence model. In each iteration,
the algorithm performs a conventional beam search but is only allowed to output a rewrite (non-identity translation) if it has high confidence i.e., its cost is less than the cost of the identity translation times a prespecified threshold. Using iterative decoding allows a stricter threshold value than what is optimal for single-shot decoding, as a change ignored for being low confidence in one decoding iteration may be selected in the next.

Using incremental edits produces a significant improvement
in performance over single-shot decoding for models trained on the Wikipedia revision data, a highly noisy corpus, while models trained on the relatively clean round-trip translation data see no improvment. All models finetuned on Lang-8 see improvement with iterative decoding (Figure~\ref{fig:iterative}, Table~\ref{tab:singleshotiterative}).

\RestyleAlgo{boxruled}

\begin{algorithm}
\small
\KwData{$I$, $\text{beam}$, $\text{threshold}$, $\text{MAXITER}$}
\KwResult{$\hat{T}$}
\For{$i \in \{1,2,...,\text{MAXITER}\}$} {
  $\text{Nbestlist} = \text{Decode}(I, \text{beam})$ \\
  $C_{\text{Identity}} = +\infty$ \\
  $C_{\text{Non-Identity}} = +\infty$ \\
  $H_{\text{Non-Identity}} = NULL $ \\
  \For{$H \in \text{Nbestlist}$} {
    \uIf{$H = I$} {
      $C_{\text{Identity}} = \text{Cost}(H)$; \\
    }
    \uElseIf {\text{Cost}(H) $<$ $C_\text{Non-Identity}$} {
      $C_{\text{Non-Identity}} = \text{Cost}(H)$ \\
      $H_{\text{Non-Identity}} = H$
    }
  }
  \Comment{Rewrite if non-identity cost $<$ identity cost}\\
  \eIf{$C_{\text{Non-Identity}} / C_{\text{Identity}} < \text{threshold}$} {
    $\hat{T} = H_{\text{Non-Identity}}$ \Comment{Output rewrite.}
  }{
    $\hat{T} = I$ \Comment{Output identity.}
  }
  $I = \hat{T}$ \Comment{Input for next iteration.}
}
\caption{Iterative Decoding}
\label{algo:decode}
\end{algorithm}

In Table~\ref{tab:decode}, we show an example of iterative decoding in
action. The model continues to refine the input until it reaches a
sentence that does not require any edits. We generally see fewer edits
being applied as the model gets closer to the final result.

\section{Model}
\label{sec:model}
In this work, we use the \textit{Transformer} sequence-to-sequence model~\cite{vaswani2017attention}, using the \textit{Tensor2Tensor} opensource
implementation.\footnote{\url{https://github.com/tensorflow/tensor2tensor}} We use 6 layers for both the encoder and the decoder, 8 attention heads, embedding size $d_{\text{model}} = 1024$, a position-wise feed forward network at every layer of inner size $d_{ff} = 4096$, and Adafactor as optimizer with inverse squared root decay~\cite{shazeer2018adafactor}\footnote{We used the ``transformer\_clean\_big\_tpu'' setting.}. The word tokens are split into subwords using a variant of the byte-pair encoding technique~\cite{sennrich2016b}, described in ~\newcite{schuster2012wordpiece}.

We train the {\it Transformer} model  for 5 epochs with a batch size of approximately 64,000 word pieces. While training on the Wikipedia corpora, we set the learning rate to 0.01 for the first 10,000 steps, then
decrease it proportionally to the inverse square root of the number of steps after that.

We then finetune our models on Lang-8 for 50 epochs and use a constant learning rate of $3 \times 10^{-5}$. We stop the fine-tuning before the models start to overfit on a development set drawn from Lang-8.

\section{Experiments}
\label{sec:experiments}

\subsection{Evaluation}
We report results on the CoNLL-2014 test set~\cite{ng2014conll} and the JFLEG test set~\cite{napoles2017jfleg,heilman2014}. Our initial experiments with iterative decoding showed that increasing beam sizes beyond 4 did not yield improvements in performance. Thus, we report all results using a beam size of 4. Our ensemble models are obtained by decoding with 4 identical {\it Transformers} trained and finetuned separately. Ensembles of neural translation systems are typically constructed by computing the logits from each individual system and combining them using either an arithmetic average~\cite{sutskever14} or a geometric average~\cite{cromieres16}. Similar to ~\newcite{cromieres16}, we find that a geometric average outperforms an arithmetic average. Hence, we report results using only this scheme.

Following~\cite{grundkiewicz2018near,junczys2018approaching}, we
preprocess JFLEG development and test sets with a spell-checking component but do not apply spelling correction to CoNLL sets. For CoNLL sets, we pick the best iterative
decoding threshold and number of iterations on a subset of the CoNLL-2014 training set, sampled to have the same ratio of modified to unmodified sentences as the
CoNLL-2014 dev set. For JFLEG, we pick the best decoding threshold on the JFLEG dev set.We report performance of our models by measuring \ffive with the $M^2$ scorer~\cite{dahlmeier2012better} on the CoNLL-2014 dev and test sets, and the GLEU+ metric~\cite{napoles2016gleu} on the JFLEG dev and test sets. Table~\ref{tab:corpusstats} reports statistics computed over the development and test sets.
\begin{table}[ht]
  \centering
  \footnotesize
  \begin{tabular}{c|c|c|c} \toprule
    Test/Dev Set & \# sentences & \# annotators & Metric \\ \hline
    CoNLL-2014 dev & 1345 & 1 & $M^2$ \\
    CoNLL-2014 test & 1312 & 2 & $M^2$ \\
    JFLEG dev & 754 & 4 & GLEU \\
    JFLEG test & 747 & 4 & GLEU \\ \bottomrule
  \end{tabular}
  \caption{Statistics for test/dev data.}
  \label{tab:corpusstats}
\end{table}

\subsection{Data from Wikipedia Revisions}
\begin{table}[ht]
    \centering
    \footnotesize
    \begin{tabular}{c|ccc|c}
      \toprule
      Revision Dataset & \multicolumn{3}{c}{CoNLL-2014} & JFLEG \\ \hline
    & Prec. & Rec. & \ffive & \gleu \\ \midrule
        \multicolumn{5}{c}{Revisions} \\ \hline
        % \multicolumn{5}{c}{pretrained} \\ \hline
        \textit{Default setting} & 62.7 & 24.3 & 47.7 & 56.9 \\
        \textit{Max-edit-28} & 57.3 & 28.0 & 47.4 & 57.1 \\
        \textit{Max-edit-6} & 58.3 & 25.7 & 46.5 & 56.1 \\
        \textit{Dwnsample-1.35} & 47.0 & 35.1 & 44.0 & 56.4 \\
        All & 58.3 & 25.1 & 46.1 & 56.8 \\
        \midrule
        \multicolumn{5}{c}{+ finetuning on Lang-8} \\ \hline
        \textit{Default setting} & 68.8 & 32.3 & 56.1 & 61.7 \\ % nd
        \textit{Max-edit-28} & 59.6 & 40.9 & 54.6 & 61.8 \\ % filter 28
        \textit{Max-edit-6} & 65.5 & 37.1 & 56.8 & 61.6 \\ % revision skip 1.35
        \textit{Dwnsample-1.35} & 62.7 & 39.9 & 56.3 & 61.3 \\ % filter 6
        All & 64.5 & 36.2 & 55.8 & 62.0 \\
        \midrule
        Lang-8 only & 41.2 & 16.4 & 31.7 & 52.8 \\
        \bottomrule
    \end{tabular}
    \caption{Performance of the models trained on variants of data extracted from Wikipedia revision histories (top panel) and then fine-tuned on Lang-8 (bottom panel), and of a model trained only on Lang-8 with the same architecture.}
  \label{tab:resultswikiedits}
\end{table}

In extracting examples from Wikipedia revision histories, we set a number of variables, selecting rate of revision downsampling, and maximum edit distance.
We generate four data sets using variations of these values: \textit{Default setting} uses the default values described in Section~\ref{sec:wikidata}, \textit{Max-edit-28} and \textit{Max-edit-6} correspond to maximum edit distance of 28 and 6 wordpieces respectively, and \textit{Dwnsample-1.35} corresponds to a revision downsampling rate of $\text{log}_{1.35}(n)$ for a page with a total of $n$ revisions (whereas the default setting uses a rate of $\text{log}_{1.5}(n)$). We train a fifth model on the union of the datasets. Table~\ref{tab:resultswikiedits} shows that varying the data generation parameters led to modest variation in performance, but training on the union of the diverse datasets did not yield any benefit. Fine-tuning yields large improvements for all models. As a sanity check, we also trained a model only on Lang-8 with the same architecture. All pre-trained and fine-tuned models substantially outperform this Lang-8 only model, confirming the usefulness of pre-training.

\subsection{Round Trip Translations}
As for the Revision data, we train a model on each of the round-trip translation datasets, and a fifth model on the union of their data, then fine-tune all models. The results are shown in Table~\ref{tab:rtsingle}. Using Japanese as the bridge language gives the best performance on CoNLL-2014, even when compared to the model trained on all round-trip data. This is likely because the error patterns generated using Japanese round-trip translations are very similar to those in CoNLL-2014 set, created from non-native speakers of English~\cite{ng2014conll}. Pooling all round-trip translations dilutes this similarity and lowers performance on CoNLL-2014. However, the model trained on all data performs best on the JFLEG set, which has a different distribution of errors relative to CoNLL-2014~\cite{napoles2017jfleg}. After fine-tuning, all round-trip models perform considerably better than the Lang-8 model.

\begin{table}[ht]
  \small
  \centering
  \begin{tabular}{c|ccc|c}
   \toprule
   Bridge Language & \multicolumn{3}{c}{CoNLL-2014} & JFLEG \\
   \hline
   & Precision & Recall & \ffive & \gleu \\
   \midrule
   \multicolumn{5}{c}{Round-Trip Translations} \\ \midrule
    French & 33.6 & 21.9 & 30.3 & 50.6 \\
    German & 36.4 & 21.2 & 31.8 & 51.3 \\
    Russian & 33.5 & 21.1 & 30.0 & 50.5 \\
    Japanese & 35.7 & 51.3 & 38.1 & 46.2 \\
    All & 38.1 & 27.1 & 35.2 & 52.1 \\
    \midrule
   %% TODO(shankarkumar): Update the tables below : FT/Ensemble
   \multicolumn{5}{c}{+ finetuning on Lang-8} \\ \hline
    French & 57.9 & 39.9 & 53.1 & 60.9 \\
    German & 56.4 & 42.1 & 52.8 & 61.5 \\
    Russian & 60.8 & 32.5 & 51.7 & 59.9 \\
    Japanese & 60.9 & 38.6 & 54.6 & 61.5 \\
    All & 62.1 & 40.0 & 56.0 & 61.6 \\
    \midrule
    Lang-8 only & 41.2 & 16.4 & 31.7 & 52.8 \\
    \bottomrule
  \end{tabular}
  \caption{Performance of the models trained on the round-trip translations (top panel) and fine-tuned on Lang8 (bottom panel) and of a model trained only on Lang-8 with the same architecture.}
  \label{tab:rtsingle}
\end{table}

\subsection{Combining Data Sources}
Having generated multiple diverse datasets, we investigate strategies for utilizing combinations of data from multiple sources. For each corpus, we train a single model on all data and compare its performance to an ensemble of the 4 individually-trained models (Table~\ref{tab:roundtripensemble}). The ensemble clearly outperforms the single model for both types of data. We additionally train a single model on the union of all Revisions and Round-Trip Translated datasets reported on in Tables~\ref{tab:resultswikiedits} and~\ref{tab:rtsingle}, which we compare to an ensemble of the 8 models trained individually on those datasets.

When Wikipedia edits are combined with the round-trip translations, the single-model performance remains unchanged on CoNLL-2014, while the ensemble shows an improvement. This suggests that when utilizing disparate sources of data, an ensemble is preferable to combining the data.

\begin{table}[h]
  \centering
  \small
  \begin{tabular}{c|ccc|c}
   \toprule
    Model & \multicolumn{3}{c}{CoNLL-2014} & JFLEG \\ \hline
    & Precision & Recall & \ffive & \gleu \\ \midrule
    \multicolumn{5}{c}{Revisions} \\ \midrule
    All & 64.5 & 36.2 & 55.8 & 62.0 \\
    Ensemble (4) & 66.3 & 42.3 & 59.0 & 62.9 \\ \midrule
    \multicolumn{5}{c}{Round-Trip Translations} \\ \midrule
    All & 62.1 & 40.0 & 56.0 & 61.6 \\
    Ensemble (4) &  63.5 & 47.0 & 59.3 & 63.2  \\ \midrule
    \multicolumn{5}{c}{Revisions + Round-Trip Translations} \\ \midrule
    All &  65.8 & 35.2 & 56.1 & 62.6 \\
    Ensemble (8) & 66.7 & 43.9 & 60.4 & 63.3 \\
    \bottomrule
  \end{tabular}
  \caption{Combining datasets using either a single model trained on all data versus an ensemble of models. All models are fine-tuned on Lang-8.}
  \label{tab:roundtripensemble}
\end{table}

\begin{table*}[ht]
    \centering
    \footnotesize
    \begin{tabular}{c|c|rrr|r}
    \toprule
        & Model & \multicolumn{3}{c|}{CoNLL-2014} & JFLEG \\
        & & Precision & Recall & \ffive & {\gleu} \\
    \midrule
    {\scriptsize \newcite{chollampatt2018multilayer}} & MLConv$_{embed}$
    % \newcite{chollampatt2018multilayer} & MLConv$_{embed}$
                                            & 60.9 & 23.7 & 46.4  % conll-test
                                            & 51.3 \\       % jfleg
    & Ensemble (4) +EO +LM +SpellCheck
                                            & 65.5 & 33.1 & 54.8  % conll-test
                                            & 57.5 \\       % jfleg
    \midrule
    {\scriptsize \newcite{junczys2018approaching}} & Single Transformer
    % \newcite{junczys2018approaching} & Single Transformer
                                            & & & 53.0     % conll-test
                                            & 57.9 \\  % jfleg
        & Ensemble (4)
                                            & 63.0 & 38.9 & 56.1  % conll-test
                                            & 58.5 \\           % jfleg
        & Ensemble (4) +LM
                                            & 61.9 & 40.2 & 55.8  % conll-test
                                            & 59.9 \\          % jfleg
    \midrule
    {\scriptsize \newcite{grundkiewicz2018near}} & Hybrid PBMT +NMT +LM
    % \newcite{grundkiewicz2018near} & Hybrid PBMT +NMT +LM
                                            & 66.8 & 34.5 & 56.3  % conll-test
                                            & 61.5 \\    % jfleg
    \bottomrule
    \toprule
    \multicolumn{2}{c|}{Best Single Model} & 65.5 & 37.1 & 56.8 & 61.6  \\
    \multicolumn{2}{c|}{Best Ensemble} & 66.7 & 43.9 & \textbf{60.4} 
                                & \textbf{63.3} \\
    \bottomrule
    \end{tabular}
    \caption{Comparison of recent state-of-the-art models (top) and our best single-system and ensemble models (bottom) on the CoNLL-2014 and JFLEG datsets. Only systems trained with publicly available Lang-8 and CoNLL datasets are reported.
    }
    \label{tab:results}
\end{table*}

\subsection{Comparison with Other systems}
We compare the performance of our best individual system, trained on all revisions, the best ensemble of 8 models trained from both revisions and roundtrip translations on the CoNLL-2014 and JFLEG datasets (Table~\ref{tab:results}). We only report performance of models that use publicly available Lang-8 and CoNLL datasets. Our single system trained on all revisions outperforms all previous systems on both datasets, and our ensemble improves upon the single system result\footnote{Using non-public sentences beyond the regular Lang-8 and CoNLL datasets, ~\newcite{ge2018} recently obtained an \ffive of 61.3 on CoNLL-2014 and a GLEU of 62.4 on JFLEG. Using finetuning data beyond the standard datasets, we obtain an \ffive of 62.8 on CoNLL-2014 and a GLEU of 65.0 on JFLEG.}.

\section{Error Analysis}\label{sec:eval_analysis}
\begin{table*}[ht]
  \footnotesize
  \centering
  \begin{tabular}{p{2.2cm}|p{13cm}}
    \toprule
    Original & Recently, a \textbf{new coming} surveillance technology called radio-frequency identification \textbf{which is RFID for short} has caused heated discussions on whether it should be used to track people. \\
    \midrule
    Revisions & Recently, a surveillance technology called radio frequency identification \textbf{(RFID)} has caused heated discussions on whether it should be used to track people. \\
    +finetuning & Recently, a \textbf{new} surveillance technology called radio-frequency identification\textbf{, which is RFID for short,} has caused heated discussions on whether it should be used to track people. \\
    Ensemble & Recently, a new coming surveillance technology called radio-frequency identification, which is RFID for short\textbf{,} has caused heated discussions on whether it should be used to track people. \\
    \midrule
    Round-Trip & Recently, a new coming surveillance technology called radio-frequency identification which is RFID for short has caused heated discussions on whether it should be used to track people. \\
    +finetuning & Recently, a \textbf{new upcoming} surveillance technology called radio-frequency identification which is RFID for short has caused heated discussions on whether it should be used to track people. \\
    Ensemble & Recently, a \textbf{new} surveillance technology called radio-frequency identification which is RFID for short has caused heated discussions on whether it should be used to track people. \\
  \midrule
    \midrule
    Original & \textbf{Then we can see that} the rising life \textbf{expectancies} can also be viewed as a challenge for \textbf{us} to face. \\
    Revisions & \textbf{The} rising life \textbf{expectancy} can also be viewed as a challenge for \textbf{people} to face. \\
    +finetuning & \textbf{Then we can see that} the rising life \textbf{expectancy} can also be viewed as a challenge for \textbf{us} to face. \\
    Ensemble & Then we can see that the rising life expectancies can also be viewed as a challenge for us to face. \\
    \midrule
    Round-Trip & Then we can see that the rising life expectancies can also be viewed as a challenge for us to face. \\
    +finetuning & Then we can see that the rising life \textbf{expectancy} can also be viewed as a challenge for us to face. \\
    Ensemble & Then we can see that the rising life expectancies can also be viewed as a challenge for us to face. \\
  \bottomrule
  \end{tabular}
  \caption{Corrections from models trained on (a) Wikipedia revisions and (b) round-trip translations using Japanese as a bridge language, along with suggestions from their Lang-8 finetuned counterparts. Also shown are the corrections from the ensembles of 4 wikipedia models as well as 4 models trained on round trip translations. Example sentences are from the CoNLL-2014 dev set. }
\label{tab:examples}
\end{table*}
All models trained on Wikipedia-derived data are demonstrated to benefit significantly from fine-tuning on Lang-8 (Tables~\ref{tab:resultswikiedits} and~\ref{tab:rtsingle}). In Table~\ref{tab:examples}, we compare example corrections proposed by
two Wikipedia-derived models to the corrections proposed by their fine-tuned counterparts.
The changes proposed by the revisions-trained model
often appear to be improvements to the original sentence, but fall outside the scope of GEC.
Models finetuned on Lang-8 learn to make more conservative
corrections.

The finetuning on Lang-8 can be viewed as an adaptation
technique that shifts the model from the Wikipedia-editing task
to the GEC task. On Wikipedia, it is common to see
substantial edits that make the text more concise and readable,
e.g. replacing ``which is RFID for short'' with ``(RFID)'', or
removing less important clauses like ``Then we can see that''. But
these are not appropriate for GEC as they are editorial style fixes
rather than grammatical fixes. The models trained on round-trip translation seem to be make fewer drastic changes.

\begin{table*}[ht]
  \small
  \centering
  \begin{tabular}{c|ccc|ccc} \toprule
    Error Type & \multicolumn{3}{c}{Revisions} & \multicolumn{3}{c}{Round-trip Translations} \\ \midrule
    & Pre-trained & Fine-tuned & Ensemble & Pre-trained & Fine-tuned & Ensemble \\ \midrule
    Adjective & 16.9 & 29.4 & 36.6 & 14.4 & 27.8 & 37.9 \\
    Adverb & 31.5 & 39.7 & 43.5 & 21.7 & 33.3 & 44.6 \\
    Determiner & 31.3 & 57.2 & 59.4 & 27.4 & 57.7 & 59.5 \\
    Morphology & 64.5 & 66.1 & 66.1 & 38.7 & 59.3 & 62.0 \\
    Noun & 24.1 & 28.6 & 33.2 & 8.6 & 27.5 & 32.4 \\
    Orthography & 69.4 & 57.1 & 69.6 & 19.2 & 58.6 & 57.9 \\
    Preposition & 33.0 & 49.2 & 55.6 & 30.3 & 52.7 & 61.9 \\
    Pronoun & 34.9 & 34.1 & 44.6 & 24.4 & 41.7 & 50.1 \\
    Punctuation & 26.7 & 29.5 & 36.4 & 29.8 & 18.4 & 33.3 \\
    Spelling & 60.6 & 69.2 & 66.7 & 51.0 & 58.5 & 62.5 \\
    Verb & 36.1 & 47.1 & 43.2 & 20.7 & 45.2 & 43.2 \\
    Word Order & 45.5 & 33.3 & 52.1 & 34.8 & 42.9 & 45.5 \\ \bottomrule
  \end{tabular}
  \caption{\ffive across error categories on the CoNLL-2014 test set.}
  \label{tab:errorstats}
 \end{table*}

Table~\ref{tab:errorstats} reports \ffive across broad error categories for models trained from revisions and round-trip translations on the CoNLL-2014 test set. The error categories were tagged using the approach in ~\newcite{bryant17}. Although the overall \ffive of the 2 ensembles are similar, there are notable differences on specific categories. The ensemble using round-trip translation performs considerably better on prepositions and pronouns while the revision ensemble is better on morphology and orthography. Thus, each system may have advantages on specific domains.
% \subsection{Typical Round-Trip Translation Errors}

\section{Related Work}
\label{sec:related}
Progress in GEC has accelerated rapidly since the CoNLL-2014 Shared Task~\cite{ng2014conll}. \newcite{rozovskaya2016grammatical} combined a Phrase Based Machine Translation (PBMT) model trained on the Lang-8 dataset \cite{mizumoto2011mining} with error specific classifiers. \newcite{junczys2016phrase} combined a PBMT model with bitext features and a larger language model. The first Neural Machine Translation (NMT) model to reach the state of the art on CoNLL-2014~\cite{chollampatt2018multilayer} used an ensemble of four convolutional sequence-to-sequence models followed by rescoring. The current state of the art (\ffive of 56.25 on CoNLL-2014) using publicly available Lang-8 and CoNLL data was achieved by \newcite{grundkiewicz2018near} with a hybrid PBMT-NMT system. A neural-only result with an \ffive of 56.1 on CoNLL-2014 was reported by \newcite{junczys2018approaching} using an ensemble of neural {\it Transformer} models~\cite{vaswani2017attention}, where the decoder side of each model is pretrained as a language model. From a modeling perspective, our approach can be viewed as a direct extension of this last work. Rather than pretraining only the decoder as a language model, we pretrain on a large amount of parallel data from either Wikipedia revision histories or from round-trip translations. While pretraining on out-of-domain data has been employed previously for neural machine translation~\cite{luong2015pretrain}, it has not been presented in GEC thus far, perhaps due to the absence of such large datasets. ~\newcite{ge2018} apply iterative decoding, where two neural models, trained in left-to-right and right-to-left directions, are applied in an interleaved manner. Similar to their study, we find that iterative decoding can improve the performance of GEC.

Prior work~\cite{brockett2006correcting,foster2009,rozovskaya2010},~\cite{felice2014,xie2016neural,rei2017} has investigated multiple strategies for generating artificial errors in GEC. ~\newcite{cahill2013} show that preposition corrections extracted from Wikipedia revisions improve the quality of a GEC model for correcting preposition errors. Back-translation~\cite{sennrich2016improving,xie2018backtranslate} addresses data sparsity by introducing noise into a clean corpus using a translation model trained in the clean to noisy direction. However, training such a reverse translation model also requires access to parallel data which is scarce for GEC. In contrast, round-trip translation attempts to introduce noise via bridge translations. Round-trip translations have been investigated for GEC. Madnani et al. ~\newcite{madnani2012} combine round-trip translations to generate a lattice from which the best correction is extracted using a language model. D\'esilets et al. ~\shortcite{desilets2009} use round-trip translations for correcting preposition errors. In contrast to these approaches, we employ round-trip translations for generating a large parallel training corpus for neural GEC models.

\section{Discussion}
\label{sec:discussion}
Motivated by data scarcity for the GEC task, we present two contrasting approaches for generating large parallel corpora from the same publicly available data source. We believe both techniques offer promising research avenues for further development on the task.

We show that models trained exclusively on minimally filtered English Wikipedia revisions can already be valuable for the GEC task. This approach can be easily extended to the many other languages represented in Wikipedia, presenting an opportunity to extend GEC into languages that may have no extant GEC corpora. While we expect pre-training on Wikipedia to give us a reasonable model, it may be crucial to fine-tune this model on small amounts of clean, in-domain corpora to achieve good performance.

When extracting examples from the Wikipedia revisions, we implemented minimal filtration in pursuit of simplicity, and to produce a sufficiently large dataset. Implementing more complex filtration in order to reduce the noise in the generated dataset will likely be a productive avenue to increase the value of this approach. The performance achieved by the reported Wikipedia revisions-trained models, both with and without finetuning, may be used as a baseline by which to evaluate smaller, cleaner datasets drawn from Wikipedia revisions.

Round-trip translation takes advantage of the advanced state of the task of Machine Translation relative to GEC by leveraging extant translation models as a source of grammatical-style data corruption. In this work, we only experiment with producing English-language GEC corpora, but this technique can be extended to any of the many languages for which translation models exist. It would be useful to assess how the translation quality influences the performance of the resulting GEC model. In our experiments with round-trip translation, we used target sentences drawn from Wikipedia to maintain a reasonable comparability between the two techniques. However, there is no constraint preventing the application of round-trip translation to diverse data sources; any source of clean text can be turned into a parallel GEC corpus. This can be used to increase diversity in the generated data, or to generate domain-specific GEC corpora (e.g. patents).

We observe that pooling two diverse data sources used to train competitively performing models on the same task can degrade performance. This suggests that within datasets useful for a specific task, there may be greater value to be discovered in finding optimal partitions of the data for training models which can then be combined using ensembles. Prior work in combining diverse data sources includes addition of special tokens~\cite{johnson17} and meta-learning~\cite{finn17}. We intend to compare ensembling with these alternatives.

We have opensourced the scripts used to extract example pairs from Wikipedia, which we hope will become a resource in the further development of models for GEC as well as other NLP tasks that rely on edit histories, such as sentence re-writing~\cite{faruqui18} and text simplification~\cite{tonelli16}.\\
\\ \textbf{Acknowledgements} \\ \\
We thank Jayakumar Hoskere, Emily Pitler, Slav Petrov, Daniel Andor, Alla Rozovskaya and Antonis Anastasopoulos for helpful suggestions. We also thank Jayakumar Hoskere, Shruti Gupta and Anmol Gulati for providing various GEC resources that were used in this paper.

\bibliographystyle{acl_natbib}
\bibliography{tgec}

\appendix

\end{document}